\documentclass{vgtc}                 

\graphicspath{{figures/}{pictures/}{images/}{./}} 
\usepackage{amsmath}
\usepackage{times}                     

\usepackage{tabu}                      
\usepackage{booktabs}                  
\usepackage{lipsum}                    
\usepackage{mwe}    
\usepackage{amssymb}

\usepackage{mathptmx}                  

\onlineid{0}

\vgtccategory{Research}

\vgtcinsertpkg

\title{Towards Cybersickness Severity Classification from VR Gameplay Videos Using Transfer Learning and Temporal Modeling}

\author{Jyotirmay Nag Setu\thanks{e-mail: jyotirmaynag.setu@utsa.edu}\\ %
        \scriptsize The University of Texas at San Antonio %
\and Kevin Desai\thanks{e-mail: kevin.desai@utsa.edu}\\ %
     \scriptsize The University of Texas at San Antonio %
\and John Quarles\thanks{e-mail: john.quarles@utsa.edu}\\ %
     \scriptsize The University of Texas at San Antonio %
     }

\teaser{
  \centering
  \includegraphics[width=\linewidth]{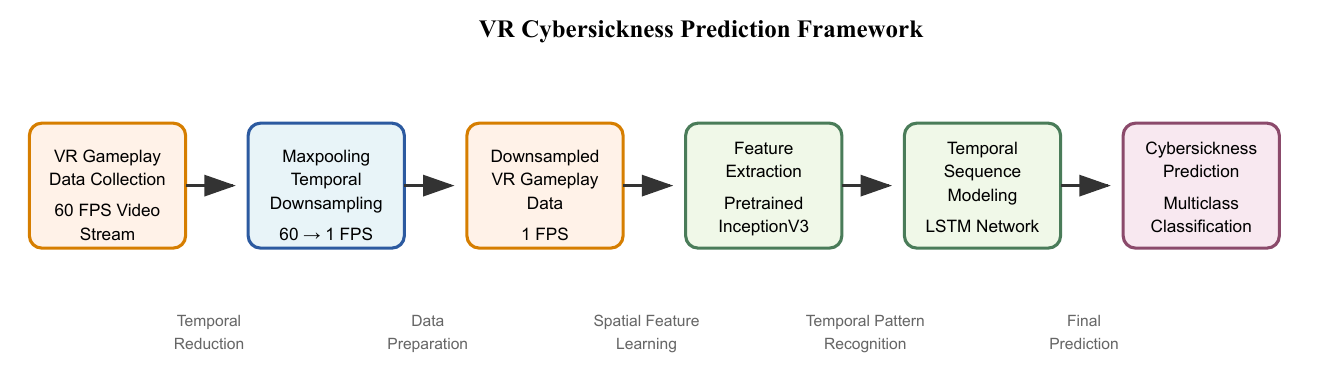}
  \caption{Overview of Transfer Learning Based Cybersickness Prediction}
  \label{fig:teaser}
}

\abstract{
    With the rapid advancement of virtual reality (VR) technology, its adoption across domains such as healthcare, education, and entertainment has grown significantly. However, the persistent issue of cybersickness, marked by symptoms resembling motion sickness continues to hinder widespread acceptance of VR. While recent research has explored multimodal deep learning approaches leveraging data from integrated VR sensors like eye and head tracking, there remains limited investigation into the use of video-based features for predicting cybersickness. In this study, we address this gap by utilizing transfer learning to extract high-level visual features from VR gameplay videos using the InceptionV3 model pretrained on the ImageNet dataset. These features are then passed to a Long Short-Term Memory (LSTM) network to capture the temporal dynamics of the VR experience and predict cybersickness severity over time. Our approach effectively leverages the time-series nature of video data, achieving an 68.4\% classification accuracy for cybersickness severity. This surpasses the performance of existing models trained solely on video data, providing a practical tool for VR developers to evaluate and mitigate cybersickness in virtual environments. Furthermore, this work lays the foundation for future research on video-based temporal modeling for enhancing user comfort in VR applications.
} 


\begin{document}


\firstsection{Introduction}

\maketitle

With the increasing availability of consumer-grade Virtual Reality (VR) devices, VR applications have expanded across diverse domains including gaming, medical training, therapy, education, and sports. VR provides fully immersive and interactive environments \cite{brooks1999s}, but this immersion is often accompanied by cybersickness\cite{davis2014systematic}. Studies report that between 61\% and 80\% of users experience cybersickness \cite{lawson2014motion}, making it a significant barrier to the widespread adoption of VR, particularly in high-stakes settings like rehabilitation and training.
While prior research has utilized bio-physiological signals (e.g., ECG, heart rate, GSR) to predict cybersickness \cite{jasper2023predicting, tasnim2024investigating, chang2021predicting, islam2022towards, islam2020automatic}, such approaches often require external sensors and limit user movement \cite{kim2017measurement, kim2019deep}, making them unsuitable for dynamic VR experiences. As a result, recent efforts have shifted toward using data directly captured by head-mounted displays (HMDs), such as eye and head tracking \cite{arcioni2019postural, chang2021predicting}, or video data from gameplay sessions \cite{lee2019motion, padmanaban2018towards}. However, most prior video-based studies rely on short, pre-recorded 360-degree videos lacking interactive locomotion and manipulation, which are important elements of modern VR gameplay \cite{melo2018presence, risi2019effects}.

Furthermore, the complexity of raw VR gameplay video data poses challenges for explainability and prediction \cite{islam2021cybersickness, lee2019motion}. Previous work has attempted to mitigate this by extracting low-level features (e.g., optical flow, brightness, contrast) \cite{jin2018automatic}, but these methods may not capture the semantic richness and temporal patterns embedded in VR scenes.

In this study, we propose a novel deep learning-based pipeline for predicting cybersickness severity from VR gameplay video. Rather than using handcrafted low-level features, we employ transfer learning with InceptionV3, a deep convolutional neural network pretrained on the ImageNet dataset. ImageNet is a large-scale dataset containing over 14 million labeled images across 1,000 categories, enabling models like InceptionV3 to learn rich, general-purpose visual features. By applying InceptionV3 to individual video frames, we extract high-level semantic features that capture visual context more effectively than handcrafted descriptors. We then input the frame-level features into a Long Short-Term Memory (LSTM) network to model the temporal structure of the video and classify cybersickness severity into four levels.

We evaluate our approach on a publicly available VR dataset: VRWalking \cite{setu2024mazed}, which includes long-form VR sessions involving navigation, manipulation, and varied tasks. Our method achieves a classification accuracy of 68.4\%, significantly outperforming prior approaches using raw video data, which reported only 54\% accuracy on the SET dataset \cite{islam2021cybersickness}.

\textbf{The contributions of this paper are summarized as follows:}
\begin{itemize}
\item We propose a deep learning-based video processing pipeline that combines transfer learning with InceptionV3 and LSTM to model spatial-temporal patterns in VR gameplay for cybersickness prediction.
\item We demonstrate that high-level semantic features extracted from ImageNet-pretrained models offer superior representation power over handcrafted video features for classifying cybersickness.
\item We evaluate our model on VRWalking and achieve a 68.4\% classification accuracy across four levels of cybersickness severity, outperforming previous video-based models.
\item Our approach eliminates the need for external bio-physiological sensors, making it scalable and compatible with modern consumer VR systems.
\end{itemize}

\section{Background}

Cybersickness, a common side effect of immersive VR experiences, manifests through symptoms such as nausea, dizziness, and disorientation \cite{stanney1997cybersickness}. As VR continues to expand across domains like education, healthcare, and entertainment, reports of cybersickness have increased significantly. Although multiple theories attempt to explain its cause—most notably the sensory conflict theory \cite{laviola2000discussion}, which attributes it to mismatches between visual and vestibular inputs—its underlying mechanisms remain complex and multifaceted. Other perspectives, such as dual-task interference \cite{pashler1994dual, kasper2014isolating}, suggest cognitive load from simultaneous tasks may also contribute to discomfort in virtual environments.

Cybersickness is commonly measured through subjective instruments such as the Simulator Sickness Questionnaire (SSQ) \cite{kennedy1993simulator, balk2013simulator, bouchard2007revising}, but its low temporal resolution makes it impractical for labeling time-series data during VR sessions. To overcome this, researchers have adopted the Fast Motion Scale (FMS), a single-question instrument administered periodically during VR experiences \cite{nalivaiko2015cybersickness, nesbitt2017correlating}, often once or twice per minute \cite{setu2024mazed, islam2022towards, arshad2021reducing, dev2025investigating, setu2025predicting}.

Recent research has focused on using machine learning to predict cybersickness based on physiological signals or video data. While several studies used heart rate variability and respiration data for prediction \cite{oh2021machine, islam2020automatic}, the requirement for external sensors limits scalability and hinders free movement. Consequently, video-based prediction has gained traction as a sensor-free alternative.

Notably, Padmanaban et al.\ \cite{padmanaban2018towards} and Lee et al.\ \cite{lee2019motion} used pre-recorded stereoscopic VR videos and engineered features like depth, optical flow, and saliency to predict cybersickness, achieving RMSEs of 12.6 and 8.49, respectively. However, these approaches lacked interactivity and were limited to short exposures. Kim et al.\ \cite{kim2017measurement} trained a convolutional autoencoder using non-VR KITTI video datasets to infer motion sickness via reconstruction errors.

Jin et al.\ \cite{jin2018automatic} improved on prior work by collecting video and head tracking data from interactive VR environments, incorporating features such as brightness, contrast, and motion intensity. Their models achieved strong performance (R² = 86.8\%), though cybersickness was only measured between sessions, not during gameplay, and their study lacked explainability analysis. Islam et al.\ \cite{islam2020automatic} used 3D CNNs trained on interactive gameplay videos, achieving a classification accuracy of 54.17\% for four cybersickness severity levels, but did not fully model temporal structure or provide interpretability.

While these efforts represent important steps, they often rely on raw video data or hand-engineered features, are limited in interactivity, or lack temporal modeling. Our research builds on this foundation by applying transfer learning with InceptionV3 pretrained on ImageNet to extract high-level spatial features from interactive VR gameplay videos. These features are then fed into an LSTM network to capture the temporal evolution of the visual scene and classify cybersickness severity. This enables us to fully utilize the time-series nature of the video data while avoiding the limitations of hand-engineered or static features.
\section{VRWalking Dataset}

The VRWalking dataset, as described in \cite{setu2024mazed}, captures participants' experiences while navigating virtual mazes through physical walking over a 15-minute period. Each one-minute interval in the session was structured into two segments: the first 30 seconds involved performing an attention task, and the subsequent 30 seconds served as a Q\&A session. During the Q\&A phase, participants provided responses to Likert-scale questionnaires evaluating cybersickness, physical exertion, and mental workload, alongside reporting outcomes for a working memory task. As a result, label data was collected at one-minute intervals. This labeling strategy is consistent with methodologies adopted in prior work, which typically employ labeling windows ranging from 30 seconds to one minute \cite{wen2024vr, islam2022towards, islam2020automatic, kundu2023litevr}.

The dataset includes information from 39 participants (23 male, 16 female) with an average age of 25.67 years (SD = 7.22). Among them, 25 participants had previous VR experience, while 13 were new to VR. The sample is demographically diverse, comprising 15 Hispanic, 3 African American, 8 Asian, 11 White, 1 Alaska Native participants, and one individual who chose not to disclose their ethnicity.

Although the dataset offers a wide range of multimodal signals related to cybersickness, our study specifically focuses on the VR gameplay video recordings, which are annotated using the Fast Motion Sickness (FMS) scale.

\textit{Fast Motion Sickness Scale (FMS):} Participants were asked once per minute: “How sick do you feel on a scale from 1 (not intense) to 10 (very intense)?” \cite{keshavarz2011validating,freiwald2020cybersickness}

\textit{VR Gameplay Data:} The left-eye perspective of each participant's VR gameplay was recorded using OBS Studio at a frame rate of 60 FPS as they navigated the maze environment.

\section{Experimental Setup}

To classify cybersickness severity from VR gameplay videos, we designed a spatial-temporal deep learning pipeline that integrates high-level visual feature extraction with temporal sequence modeling. The pipeline consists of two stages: (1) frame-wise feature extraction using a pretrained InceptionV3, and (2) sequence classification using a Long Short-Term Memory (LSTM) network.

We applied 5-fold stratified cross-validation to evaluate the generalization ability of our model. Each fold consists of 80\% training data and 20\% test data, ensuring class balance across splits. The model was trained for 50 epochs using the Adam optimizer with a learning rate of 0.0001 and categorical cross-entropy loss. We used a batch size of 32 and applied early stopping if the validation accuracy did not improve for 5 consecutive epochs. Accuracy and loss were computed on the held-out test set for each fold.

\section{Feature Extraction}
\subsection{Video Downsampling Strategy}

Processing full-resolution temporal sequences from high-frame-rate VR gameplay videos can be computationally expensive and memory-intensive, especially when dealing with frame-level embeddings from deep convolutional networks. To address this challenge, we employed a temporal downsampling strategy using max pooling over fixed-size windows.

After extracting 2048-dimensional feature vectors from each frame using the pretrained InceptionV3 model, we applied 1D max pooling along the temporal axis. Let $F \in \mathbb{R}^{T \times D}$ represent the sequence of extracted frame features, where $T$ is the number of frames and $D$ is the feature dimensionality (2048 in our case). Given a pooling window size $k$, the downsampled sequence $F' \in \mathbb{R}^{\frac{T}{k} \times D}$ is computed as:

\begin{equation}
    F'_i = \max(F_{(i-1)k+1}, F_{(i-1)k+2}, \dots, F_{ik}), \quad \text{for } i = 1, 2, \dots, \left\lfloor \frac{T}{k} \right\rfloor
\end{equation}

This operation retains the most prominent activations in each window of $k$ consecutive frames, effectively preserving salient motion and visual cues while significantly reducing the sequence length. In our implementation, we used $k = 15$ to reduce a 60-frame input sequence (corresponding to 1 FPS over 60 seconds) to a 4-step LSTM input. This compressed representation balances temporal resolution and model efficiency.

Max pooling offers two key advantages in this context. First, it is robust to small variations in frame timing and minor noise, making it well-suited for naturalistic VR gameplay where scene dynamics may fluctuate. Second, by selecting the maximum activation within each window, the model emphasizes the most prominent visual features, which are likely to be more relevant for cybersickness prediction—such as sudden brightness changes, high contrast motion, or camera acceleration artifacts.

Overall, this strategy provides an efficient and effective mechanism for temporal dimension reduction without losing critical information needed for downstream sequence modeling via LSTM.

\subsection{Transfer Learning with InceptionV3}
Instead of training a visual feature extractor from scratch, we leveraged transfer learning using the InceptionV3 convolutional neural network pretrained on the ImageNet dataset. ImageNet is a large-scale image classification dataset containing over 14 million labeled images spanning 1,000 object categories. Due to its diversity and size, models pretrained on ImageNet are capable of extracting robust and general-purpose visual features.

We processed each VR gameplay video by sampling frames at a fixed rate (e.g., 1 frame per second). These frames were resized to $224 \times 224$ pixels and passed through the pretrained InceptionV3 model, with the top classification layers removed. We extracted the output from the final global average pooling layer, resulting in a 2048-dimensional feature vector per frame.

\subsection{Model Architecture}

To classify cybersickness severity from the temporally downsampled VR video features, we implemented a deep sequential neural network using two stacked Long Short-Term Memory (LSTM) layers followed by dense classification. The architecture is designed to capture both short- and long-range temporal dependencies in the high-level visual features extracted from VR gameplay.

The input to the model is a sequence of shape $(5, 10240)$, where each sequence contains 5 temporal steps (after max pooling), and each step is a concatenation of 5 consecutive InceptionV3 frame embeddings (each 2048-dimensional). Thus, each input sample is a matrix of dimension $5 \times (5 \times 2048)$.

The architecture consists of the following layers:

\begin{itemize}
    \item \textbf{First LSTM Layer:} This layer contains 100 hidden units and returns the full sequence of hidden states across the 5 time steps (\texttt{return\_sequences=True}). This allows the next LSTM layer to further model temporal dependencies.
    \item \textbf{Dropout Layer:} A dropout rate of 0.2 is applied to mitigate overfitting by randomly deactivating neurons during training.
    \item \textbf{Second LSTM Layer:} Another LSTM with 100 units processes the output sequence and returns only the final hidden state, summarizing the entire temporal sequence into a single representation.
    \item \textbf{Second Dropout Layer:} A second dropout with the same rate is applied for additional regularization.
    \item \textbf{Dense Output Layer:} A fully connected layer with 3 output units and a softmax activation function is used to produce class probabilities corresponding to the three levels of cybersickness severity.
\end{itemize}

The model is compiled using the Adam optimizer with a learning rate of 0.001. We use the \texttt{sparse\_categorical\_crossentropy} loss function, which is appropriate for multi-class classification problems where the target labels are provided as integer class indices. Model performance is evaluated using classification accuracy.

\section{Results}

We evaluated the model's performance on five test folds. The test accuracy was consistent across all folds, yielding a mean classification accuracy of 68.44\%. Table~\ref{tab:results} reports the accuracy and loss values for each fold.
\begin{figure}[h]
    \centering
    \includegraphics[width=\columnwidth]{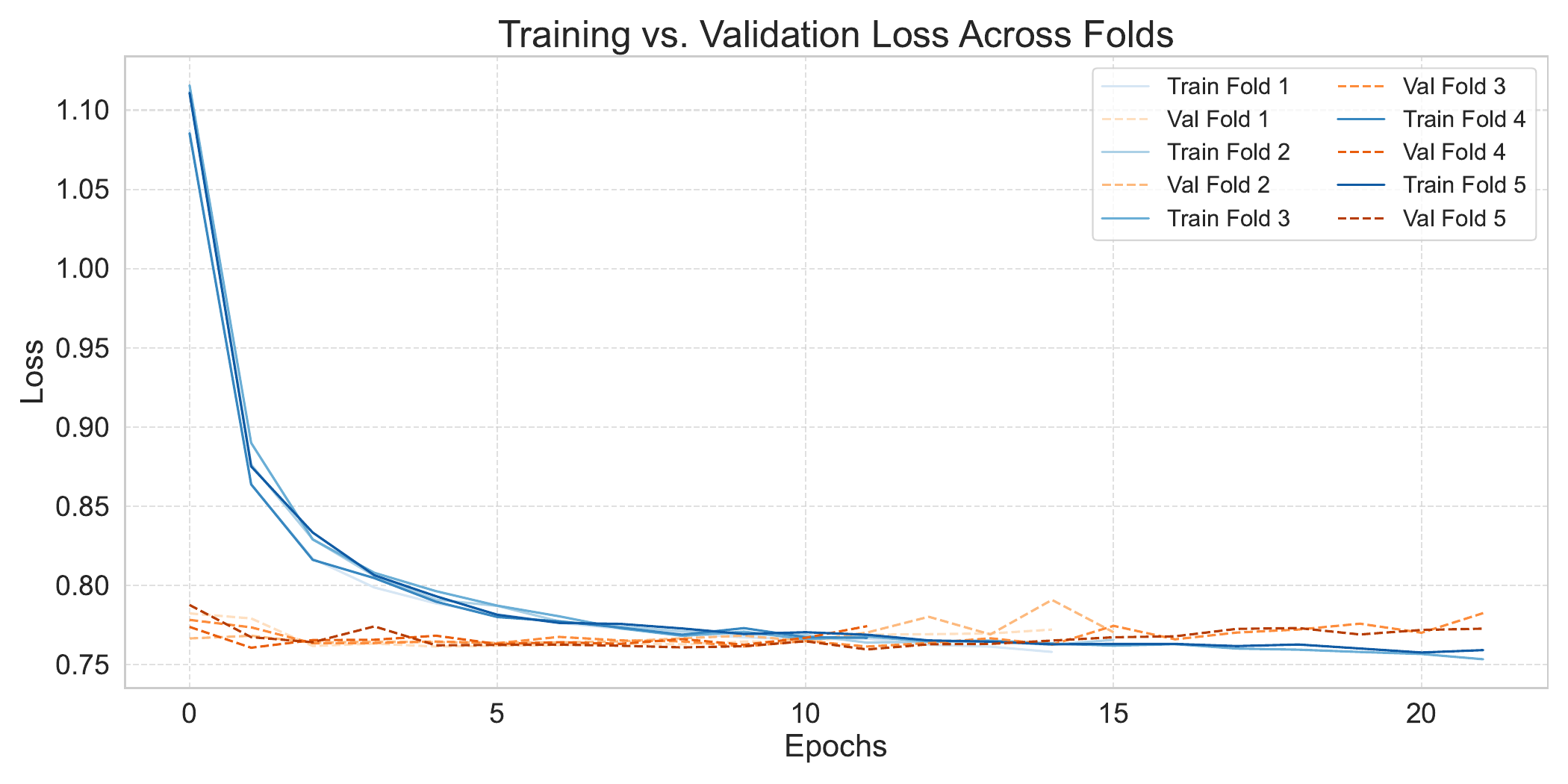}
    \caption{Training and Validation Loss over Folds}
    \label{fig:train-val-loss}
\end{figure}
\begin{table}[h]
\centering
\caption{Classification Accuracy and Loss per Fold}
\label{tab:results}
\begin{tabular}{ccc}
\hline
\textbf{Fold} & \textbf{Test Accuracy} & \textbf{Test Loss} \\
\hline
1 & 68.44\% & 0.7597 \\
2 & 68.44\% & 0.7599 \\
3 & 68.44\% & 0.7608 \\
4 & 68.44\% & 0.7593 \\
5 & 68.44\% & 0.7607 \\
\hline
\textbf{Average} & \textbf{68.44\%} & \textbf{0.7601} \\
\hline
\end{tabular}
\end{table}

The consistency of the test performance across all folds demonstrates the robustness of our transfer learning and LSTM-based architecture for modeling cybersickness in VR gameplay data. This pipeline offers a generalizable solution without requiring any external physiological sensors, making it highly scalable for real-world VR applications.
The model demonstrated consistent performance across all five folds of cross-validation, with each fold achieving an identical test accuracy of 68.44\%. This uniformity indicates that the model generalizes well and is not highly sensitive to specific data splits. Furthermore, the training and validation loss curves shown in Figure 2 confirm that the model did not overfit during training. The validation loss closely tracks the training loss across epochs in all folds, without significant divergence or instability. This stability suggests that the model maintains a balanced bias-variance tradeoff and that the features extracted from InceptionV3 provide a robust and transferable representation across participants. The combined evidence from accuracy metrics and training dynamics supports the reliability and generalization capacity of the proposed approach.
\section{Explanation of Model Prediction}
We have utilized two common methods to explain the cybersickness prediction. Although the inputs of the LSTM are deep features coming from InceptionV3, with these methods we can explain the behaviour of the used LSTM.
Here, standard gradients compute the immediate sensitivity of the model's output to small changes in each input feature - essentially asking "if I slightly modify this input, how much does the prediction change?" However, standard gradients can be noisy and may saturate in deep networks, sometimes highlighting less meaningful patterns. Integrated gradients, on the other hand, provide a more robust attribution method by computing gradients along a path from a baseline (typically zeros) to the actual input, then integrating these gradients and scaling by the input difference. This approach satisfies important mathematical properties like sensitivity and implementation invariance, making it more reliable for understanding genuine feature contributions rather than just local sensitivities.
\begin{figure}[h]
    \centering
    \includegraphics[width=\columnwidth]{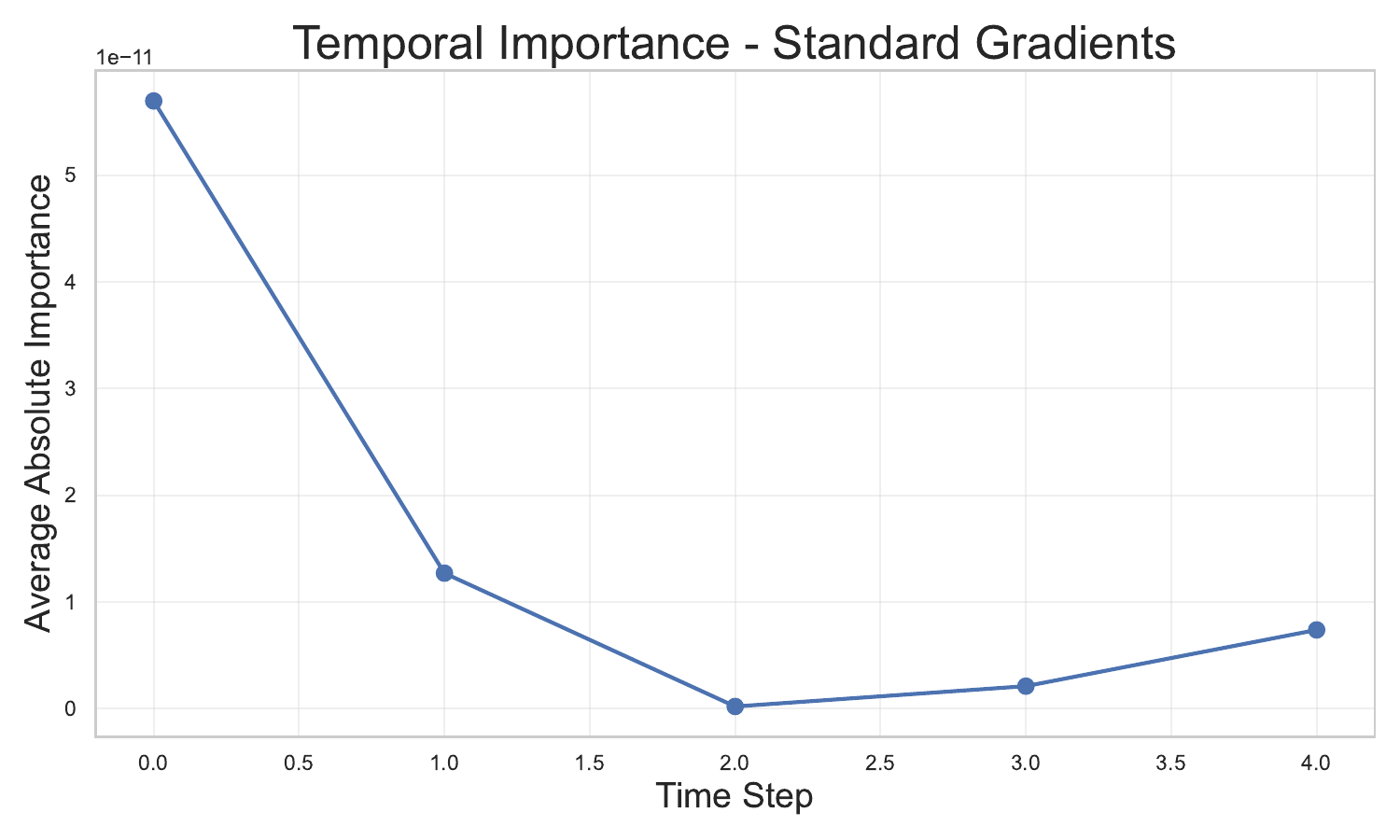}
    \caption{Temporal Importance - Standard Gradients}
    \label{fig:tem-stan}
\end{figure}
In Figure ~\ref{fig:tem-stan}, the standard gradients plot shows an "immediate onset" pattern, with the highest importance at the very beginning (time step 0) of the 5-segment sequence, then declining through segments 1 and 2, before showing a slight resurgence in segments 3 and 4. This suggests that the model's raw sensitivity is strongest to the initial visual conditions when the player first encounters this particular gameplay scenario. In the context of cybersickness, this could indicate that the model immediately detects visual patterns associated with motion sickness triggers - perhaps the onset of artificial locomotion, a sudden camera movement, or specific environmental elements that historically correlate with discomfort. However, this early peak might also reflect the inherent noisiness of standard gradients, potentially overemphasizing the starting conditions due to the method's sensitivity to local variations rather than true feature importance.
\begin{figure}[h]
    \centering
    \includegraphics[width=\columnwidth]{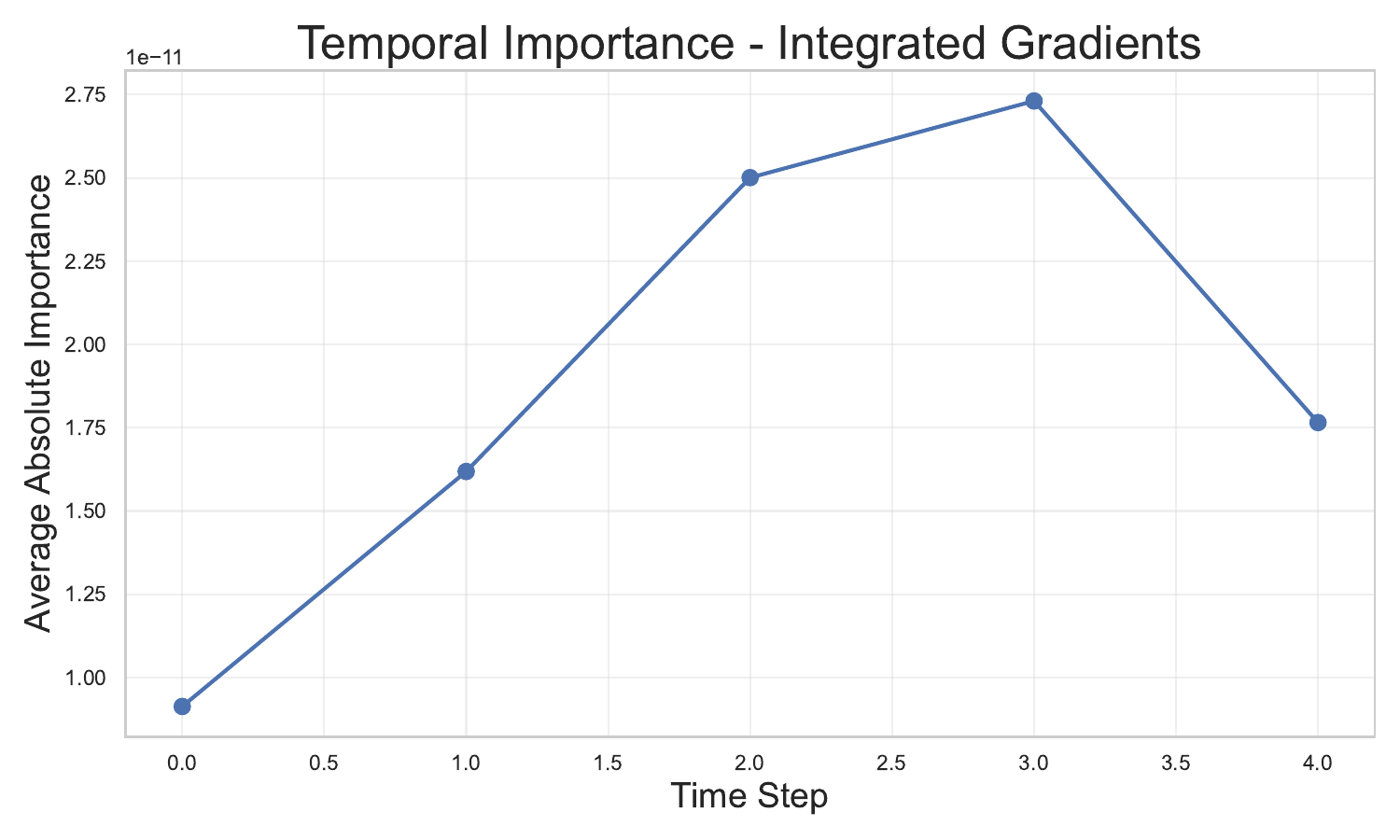}
    \caption{Temporal Importance - Integrated Gradients}
    \label{fig:tem-int}
\end{figure}
In contrast, the integrated gradients in figure ~\ref{fig:tem-int} reveal a "progressive buildup" pattern, starting low at time step 0 and steadily climbing to peak importance at segment 3, before declining in segment 4. This method, being more stable and focused on cumulative feature contributions, suggests that cybersickness symptoms in this sample develop gradually over the sequence rather than being triggered immediately. This aligns well with the physiological reality of motion sickness, where sensory conflicts often need time to accumulate before symptoms manifest. The peak at segment 3 indicates the critical moment where visual-vestibular mismatches reach a threshold, while the decline in segment 4 might represent either adaptation or the aftermath of peak symptom development. This progressive pattern is likely more trustworthy than the standard gradients due to integrated gradients' superior theoretical foundations and stability in deep networks.

However, it's important to note that interpreting these patterns comes with inherent limitations - the input features are high-dimensional abstractions from InceptionV3 (2048 features per time step), representing complex visual concepts that may not directly correspond to obvious motion patterns. These deep features capture everything from low-level optical flow and edge detection to high-level scene understanding and object recognition, making it challenging to pinpoint exactly which visual elements (smooth locomotion, head rotation, environmental complexity, or specific game mechanics) are driving the temporal importance patterns without additional analysis of the feature-level contributions. The integrated gradients' progressive buildup pattern is particularly compelling because it suggests the model has learned to recognize the temporal dynamics of how cybersickness develops in VR - not as an instantaneous trigger, but as an accumulated sensory conflict that builds over several seconds of gameplay.
\section{Discussion}
Previous studies such as Padmanaban et al. \cite{padmanaban2018towards} and Lee et al. \cite{lee2019motion} primarily utilized short, pre-recorded stereoscopic VR clips and relied on hand-engineered features like depth, disparity, and optical flow. While these studies showed promising early results, their methodologies lacked interactivity, often constrained participants to passive viewing scenarios, and did not fully exploit the temporal characteristics of video data. More recent work by Islam et al. \cite{islam2020automatic} used raw VR gameplay video with a 3D CNN, achieving 54.17\% accuracy for cybersickness classification, but did not leverage any form of pretrained semantic feature extraction or deep temporal modeling.

In contrast, our approach uses a two-stage architecture that combines the strengths of transfer learning and sequential modeling. By extracting high-level features using the InceptionV3 model pretrained on the ImageNet dataset, we circumvent the limitations of raw pixel-based learning and capitalize on a rich representation space that has proven effective across many vision domains. These semantically meaningful features are then input into an LSTM to model the temporal evolution of VR experiences, an aspect overlooked in prior studies using only static analysis or shallow temporal fusion.

The consistency of our results, achieving 68.44\% accuracy across all five folds demonstrates the robustness of this architecture. While we do not outperform some multimodal models that integrate physiological or behavioral data, our video-only pipeline offers a lightweight, scalable solution that does not require external sensors, making it highly practical for deployment on consumer-grade VR systems such as the Meta Quest or HTC Vive.

Another important contribution is our use of long-form, interactive VR sessions drawn from the VRWalking \cite{setu2024mazed} dataset, which more realistically reflect modern VR gameplay. Unlike prior work limited to short clips or seated tasks, our data includes locomotion, manipulation, and variable task difficulty, factors that can significantly influence the onset and severity of cybersickness.

\section{Limitations and Future Work}

While our proposed framework demonstrates promising results for cybersickness classification using VR gameplay video data, there are several limitations that should be addressed in future work. First, our current study is based on a relatively small participant pool, with data collected from only 20 users. Although our use of cross-validation mitigates overfitting and provides insight into the model's generalizability, the limited sample size restricts the diversity of user responses, interaction styles, and susceptibility to cybersickness. Future research should incorporate larger and more demographically varied populations to better capture individual differences and improve the robustness of predictive performance. Second, we evaluated our approach using data from a single dataset, which may limit its generalizability across different VR applications, hardware platforms, or content genres. Cybersickness can vary significantly depending on factors such as locomotion type, field of view, and scene complexity. As such, future work should evaluate the model on multiple datasets or in-the-wild scenarios to assess its cross-context reliability and adaptability. Third, our feature extraction process is based on applying a 2D image-based model (InceptionV3) to individual frames, followed by temporal modeling with an LSTM. While this approach captures coarse temporal dynamics, it may miss finer spatiotemporal patterns that are better represented using models trained directly on video sequences. Recently, pretrained video encoders such as TimeSformer, VideoMAE, and I3D have shown strong performance in video scene understanding tasks. Future work could integrate these models to extract richer motion-aware representations and improve the sensitivity of the model to motion-related cues that contribute to cybersickness.

By addressing these limitations, future research can build on our findings to develop more robust, personalized, and context-aware systems for cybersickness prediction and mitigation in VR environments.

\section{Conclusion}

In this study, we presented a deep learning-based framework for cybersickness classification using only VR gameplay video data. By combining transfer learning with InceptionV3 and temporal modeling via LSTM, our approach leverages both spatial semantics and temporal dynamics to predict cybersickness severity. Evaluated on VRWalking, our model achieved a consistent accuracy of 68.44\%, outperforming previous video-based methods while maintaining scalability and ease of deployment.

This work bridges a key gap in cybersickness prediction literature by demonstrating that semantically rich and temporally aware models can be built without requiring external sensors or intrusive instrumentation. In doing so, we provide a viable pathway toward real-time, in-situ cybersickness monitoring and mitigation in consumer VR systems. Future research could extend this approach by integrating multimodal signals, adopting self-supervised learning strategies, or applying attention-based temporal architectures to further improve predictive performance and interpretability.


\bibliographystyle{abbrv-doi}

\bibliography{template}
\end{document}